\documentclass[10pt,twocolumn,letterpaper]{article}

\usepackage[pagenumbers]{cvpr} 
\usepackage{times}
\usepackage{epsfig}
\usepackage{graphicx}
\usepackage{amsmath}
\usepackage{amssymb}
\usepackage{breqn}
\usepackage{booktabs}
\usepackage{multirow}
\usepackage{tabu}
\usepackage{bm}
\usepackage{array}
\usepackage{fontawesome}
\usepackage{verbatim}
\usepackage{makecell} 
\usepackage{algorithm}
\usepackage{listings}
\usepackage[section]{placeins}
\usepackage[accsupp]{axessibility}  
\usepackage[usenames,dvipsnames]{color}
\usepackage{colortbl}
\usepackage{subcaption}
\usepackage[table,dvipsnames]{xcolor}
\usepackage[pagebackref=true,breaklinks=true,letterpaper=true,colorlinks,bookmarks=false]{hyperref}

\definecolor{mygray}{gray}{.9}
\definecolor{LightCyan}{rgb}{0.88,1,1}
\definecolor{lightcoral}{RGB}{240,128,128}
\definecolor{lightsalmon}{RGB}{255,160,122}
\definecolor{lightgray}{RGB}{128,128,128}
\definecolor{lightblue}{RGB}{212,239,251}
\definecolor{lightgreen}{RGB}{220,255,220}
\definecolor{lightpink}{RGB}{255,182,193}
\definecolor{cornsilk}{RGB}{255,248,220}
\definecolor{mintcream}{RGB}{245,255,250}
\definecolor{lavenderblush}{RGB}{255,240,245}
\definecolor{ghostwhite}{RGB}{248,248,255}
\definecolor{honeydew}{RGB}{240,255,240}
\definecolor{aliceblue}{RGB}{240,248,255}
\definecolor{ivory}{RGB}{255,255,240}
\definecolor{azure}{RGB}{240,255,255}
\definecolor{seashell}{RGB}{255,245,238}
\definecolor{textyellow}{RGB}{255,215,0}

\usepackage[capitalize]{cleveref}
\crefname{section}{Sec.}{Secs.}
\Crefname{section}{Section}{Sections}
\Crefname{table}{Table}{Tables}
\crefname{table}{Tab.}{Tabs.}

\def\cvprPaperID{2955} 
\def\confName{ICCV}
\def\confYear{2023}

\begin{document}

\title{Realistic Full-Body Tracking from Sparse Observations via Joint-Level Modeling}

\author {
    Xiaozheng Zheng \footnotemark[1] \qquad 
    Zhuo Su         \footnotemark[1] \qquad  
    Chao Wen                         \qquad 
    Zhou Xue        \footnotemark[2] \qquad
    Xiaojie Jin  \\
    ByteDance Inc \\
}

\maketitle

{
  \renewcommand{\thefootnote}%
    {\fnsymbol{footnote}}
    \footnotetext{Project page: \url{https://zxz267.github.io/AvatarJLM}.}
}

{
  \renewcommand{\thefootnote}%
    {\fnsymbol{footnote}}
  \footnotetext[1]{Equal contribution.}
}
{
  \renewcommand{\thefootnote}%
    {\fnsymbol{footnote}}
  \footnotetext[2]{Corresponding author.}
}

\begin{abstract}
To bridge the physical and virtual worlds for rapidly developed VR/AR applications, the ability to realistically drive 3D full-body avatars is of great significance. Although real-time body tracking with only the head-mounted displays (HMDs) and hand controllers is heavily under-constrained, a carefully designed end-to-end neural network is of great potential to solve the problem by learning from large-scale motion data. To this end, we propose a two-stage framework that can obtain accurate and smooth full-body motions with the three tracking signals of head and hands only. Our framework explicitly models the joint-level features in the first stage and utilizes them as spatiotemporal tokens for alternating spatial and temporal transformer blocks to capture joint-level correlations in the second stage. Furthermore, we design a set of loss terms to constrain the task of a high degree of freedom, such that we can exploit the potential of our joint-level modeling. With extensive experiments on the AMASS motion dataset and real-captured data, we validate the effectiveness of our designs and show our proposed method can achieve more accurate and smooth motion compared to existing approaches. 
\end{abstract}

\vspace{-15pt} 
\section{Introduction}
Driving human avatars in VR/AR can help to bridge the gap between the physical and virtual worlds, and create a more natural and immersive user experience. However, in a typical capture setting, only the head and hands are tracked with Head Mounted Displays (HMD) and hand controllers. With limited inputs, driving the full-body avatar is inherently an underconstrained problem. Considerable endeavor has been dedicated to addressing the challenge of inferring full-body human pose exclusively through sparse AR/VR signals, head and hands. 
Although recent studies \cite{aliakbarian2022flag, choutas2022learning, dittadi2021full} have shown promising results, they are not suitable for real-time applications like VR body tracking.
With real-time performance in mind, Winkler \etal \cite{winkler2022questsim} use reinforcement learning for training and outperform kinematic approaches with fewer artifacts. But their method requires future frames, which introduces latency to the system. 
Most recent work AvatarPoser \cite{jiang2022avatarposer} solves the problem in a more practical way by combining transformer-based architecture and inverse-kinetic optimization, setting the benchmark on large motion capture datasets (AMASS). Despite the success of AvatarPoser across a wide variety of motion classes, we argue that a learning-based end-to-end method provides more merits in simplicity, robustness, and generalization compared with a hybrid method. 

Our key insight is that correlations between different body joints should be explicitly modeled for human pose estimation as body movements are highly structured and coordinated. Especially for the problem of estimating full-body motion from sparse observations, joint-level modeling is essential as the position and rotation of each joint can affect each other, and the overall body pose. By taking into account these correlations, we can derive more plausible full-body motion, even when observations are limited.
Therefore, we design a two-stage joint-level modeling framework to capture these dependencies between body joints for more accurate and smoother human motion. In the first stage, we explicitly model the joint-level features. Then we utilize these features as spatiotemporal tokens in the second stage for a transformer-based network to capture the joint-level dependencies for recovering full-body motions.

\begin{figure}
    \vspace{15pt} 
    \centering
    \includegraphics[width=0.47\textwidth]{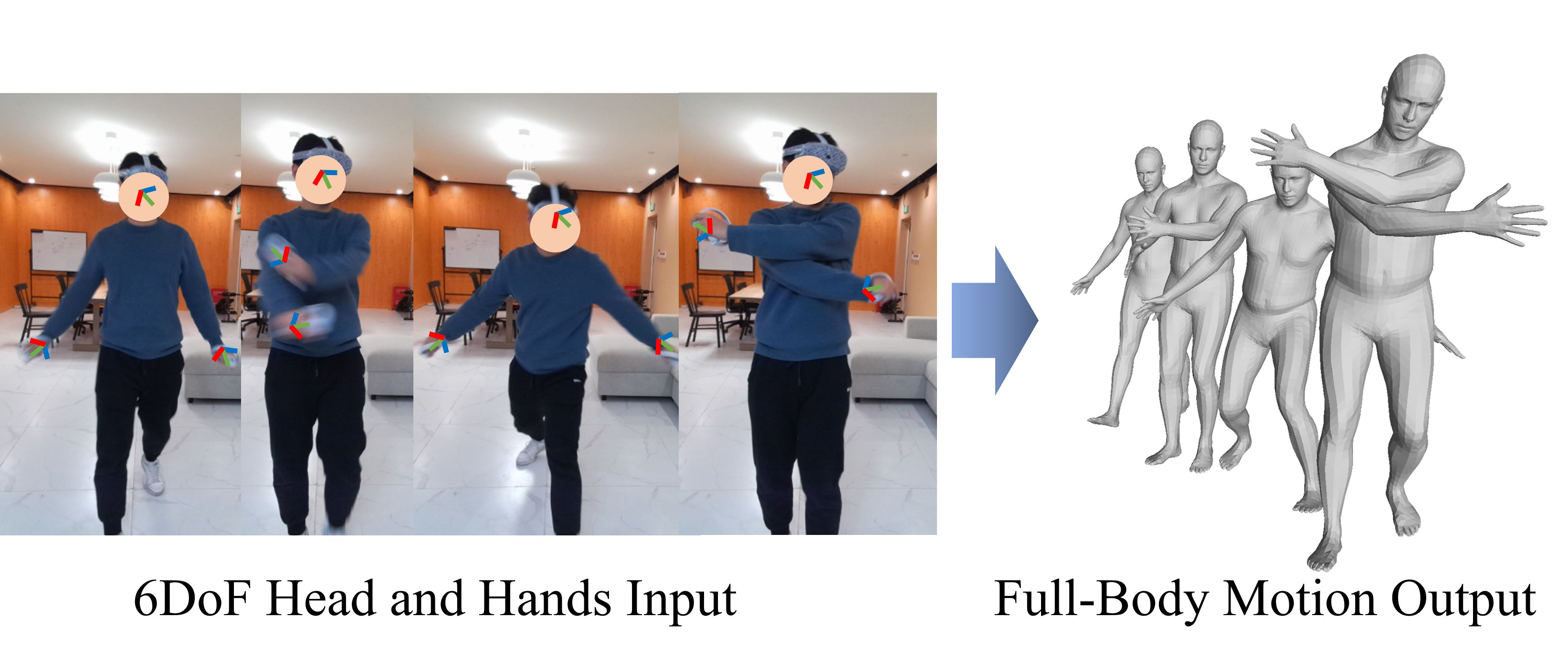}
        \vspace{-8pt} 
        \caption{Our method accurately estimates full-body motion using only head and hand tracking signals.}
    \label{fig:teaser.} 
    \vspace{-15pt} 
\end{figure}

\begin{figure*}
	\centering
	\includegraphics[width=\textwidth]{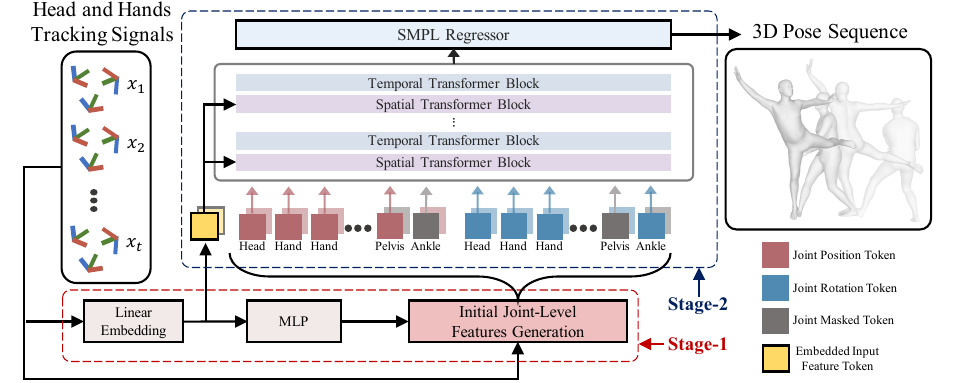}
 	\vspace{-15pt} 
    \caption{
    Illustration of our two-stage joint-level modeling framework. 
    In the first stage, we embed a sequence of sparse signals to high-dimensional input features. 
    Then, we utilize an MLP to obtain the initial full-body poses from the features. 
    After that, we combine the initial full-body poses and the sparse input signals to generate initial joint-level features. 
    In the second stage, we convert the initial joint-level features to joint-level tokens and then feed those tokens to a transformer-based network to capture the joint-level dependencies in spatial and temporal dimensions alternatively.
    In each spatial transformer block, we supplement an additional embedded input features token generated from the high-dimensional input features.
    Finally, we employ an SMPL regressor to transform the spatiotemporal modeled joint-level features into the 3D full-body pose sequence.
    }
 	\vspace{-4pt} 
	\label{fig:pipeline} 
 	\vspace{-12pt} 
\end{figure*}

In the first stage, we explicitly model the joint-level features as 1) \emph{joint-rotation features} and 2) \emph{joint-position features}. Joint-rotation offers higher compactness and computational efficiency, whereas joint-position enables more precise control and is intuitively easier to comprehend.  By combining the advantages of both features, we attain a resilient and rational human motion with the improved ability to align endpoints more accurately. 

In the second stage, we alternatively use spatial transformer blocks and temporal transformer blocks to capture the joint dependencies. Given the ambiguity inherent in our problem, we also opt to utilize the features from the first stage of each individual frame as an \emph{embedded input features} (EIF) token for every single spatial transformer block to reinforce the influence of the input sparse observations.

To capitalize on the advantages of our joint-level modeling and mitigate the risk of overfitting in the highly under-constrained problem, we have incorporated a set of loss terms into our approach. These loss terms consist of hand alignment loss, motion smoothness loss, and physical loss, each meticulously designed to enhance the efficacy and generalizability of our body-tracking system in real-world scenarios, considering the intricate and uncertain problem nature.

Extensive experiments on the large-scale motion dataset AMASS~\cite{mahmood2019amass} have demonstrated the effectiveness of our proposed designs. We also collect real-data samples for further qualitative and quantitative evaluations.
Specifically, we conduct a thorough comparison of our approach against existing methods using various protocols. 
The comparative results show that our approach significantly outperforms existing methods in all protocols by a large margin.
Moreover, our qualitative results demonstrate a significant improvement in accuracy and smoothness over the previous state-of-the-art approach, without the need for post-processing.

In summary, our contributions are the following: 
\begin{itemize}
    \item We propose a novel two-stage network that can effectively estimate full-body motion from the sparse head and hand tracking signals with high accuracy and temporal consistency. Note that our method does not need any post-processing and significantly outperforms existing state-of-the-art approaches.
    \item We elaborately design our feature extractor that generates joint-level rotational, positional, and embedded input features. These features are then utilized as spatiotemporal tokens and processed using a transformer-based network, which allows for better modeling of joint-level correlations.
    \item We introduce a set of losses that are tailored for the task of full-body motion estimation, and experimentally demonstrate the effectiveness of these losses in achieving high accuracy while avoiding undesirable artifacts such as floating, penetration, and skating.
\end{itemize}

\section{Related Work}

\subsection{Full-Body Pose from Sparse Observations}
The task of estimating the full-body pose of a human from sparse observations has garnered significant attention in the research community 
\cite{chai2005performance,von2017sparse,huang2018deep,yi2021transpose,yi2022physical,jiang2022transformer,ahuja2021coolmoves,yang2021lobstr,dittadi2021full,aliakbarian2022flag,jiang2022avatarposer,winkler2022questsim}.
Previous studies \cite{von2017sparse,huang2018deep,yi2021transpose,yi2022physical,jiang2022transformer} have relied on multiple IMU inputs to track the signals of the head, arms, pelvis and legs to capture the full body movements. 
SIP \cite{von2017sparse} demonstrates the possibility of reconstructing accurate 3D full-body poses using only 6 IMUs without any other information (e.g., video).
Sequentially, DIP \cite{huang2018deep} further uses deep learning to achieve real-time performance and better accuracy with only 6 IMUs either.
Most recently, there has been a shift towards a more AR/VR-focused setup, with some studies \cite{ahuja2021coolmoves,yang2021lobstr,dittadi2021full,aliakbarian2022flag,jiang2022avatarposer,winkler2022questsim} exploring the use of only HMD and hand controllers as input sources for even sparser observations.

Choutas \etal \cite{choutas2022learning} propose a neural optimization method to fit a parametric human body model given the observations of the head and hands.
Aliakbarian \etal's recent work \cite{aliakbarian2022flag} uses the generative model, normalizing flow, to address the under-constrained problem. Dittadi \etal \cite{dittadi2021full} take advantage of the Variational Autoencoders to synthesize the lower body and implicitly assume the fourth 3-D input by encoding all joints relative to the pelvis. 

The work QuestSim  \cite{winkler2022questsim} and AvatarPoser \cite{jiang2022avatarposer} bear the most similarities to our own methodology. QuestSim employs reinforcement learning to facilitate physics simulation. Nonetheless, the approach requires additional future information at runtime to achieve optimal results.
AvatarPoser uses a simple transformer-based network architecture to achieve accurate and smooth full-body motions, which demonstrates the feasibility of using a discriminative method for solving the task of full-body pose estimation from sparse observations.

Following the work by Jiang \etal \cite{jiang2022avatarposer}, we further investigate the potential of discriminative methods in full-body pose estimation from sparse observations. Our approach takes into account the nature of the task in the design of the framework, allowing us to leverage discriminative models for achieving more accurate and smooth full-body motions from only head and two-hand tracking signals.

\subsection{Transformer for Human Pose Estimation}
Transformer \cite{vaswani2017attention} has emerged as a popular tool for human pose estimation tasks in recent years.  Specifically, a number of studies  \cite{he2020epipolar,lin2021end,lin2021mesh,ma2022ppt} have utilized transformers to solve the task of 3D human pose estimation from RGB images. In addition, several studies \cite{zheng20213d, zhang2022mixste,li2022mhformer} have employed transformers to lift 2D human pose to the corresponding 3D human pose. Transformers have also been widely used in motion generation tasks, as seen in studies \cite{aksan2021spatio,tevet2022human,qin2022motion}.

Two works \cite{jiang2022avatarposer, jiang2022transformer} have utilized transformer-based methods to address a problem similar to that which our proposition seeks to solve. The work from Jiang \etal \cite{jiang2022transformer}  employs a transformer network and a recurrent neural network in combination to accurately estimate full-body motion using data from six IMUs. Meanwhile, Jiang \etal \cite{jiang2022avatarposer} solve the pose estimation from HMD with a combination of a transformer network followed by traditional model-based optimization.
However, both of them only treat each frame of the observed signals as a token for the temporal transformer to process. This way does not take the task nature into account and does not model the joint-level features, which makes them hard to obtain more accurate and smooth full-body motions.

Both of these methods process observed signals as sequences of temporal tokens, which are then utilized as inputs to the transformer network in a direct manner. We argue that the direct application of input signals renders the under-constrained problem more difficult to solve. In response, we explicitly model the joint-rotation and joint-position features to enable a more robust, intuitive, and comprehensive representation. Furthermore, we utilize a transformer-based network that alternatively captures spatial and temporal joint dependencies, in a manner similar to the approach taken by MixSTE \cite{zhang2022mixste}.

\begin{figure}
    \centering
    \includegraphics[width=0.47\textwidth]{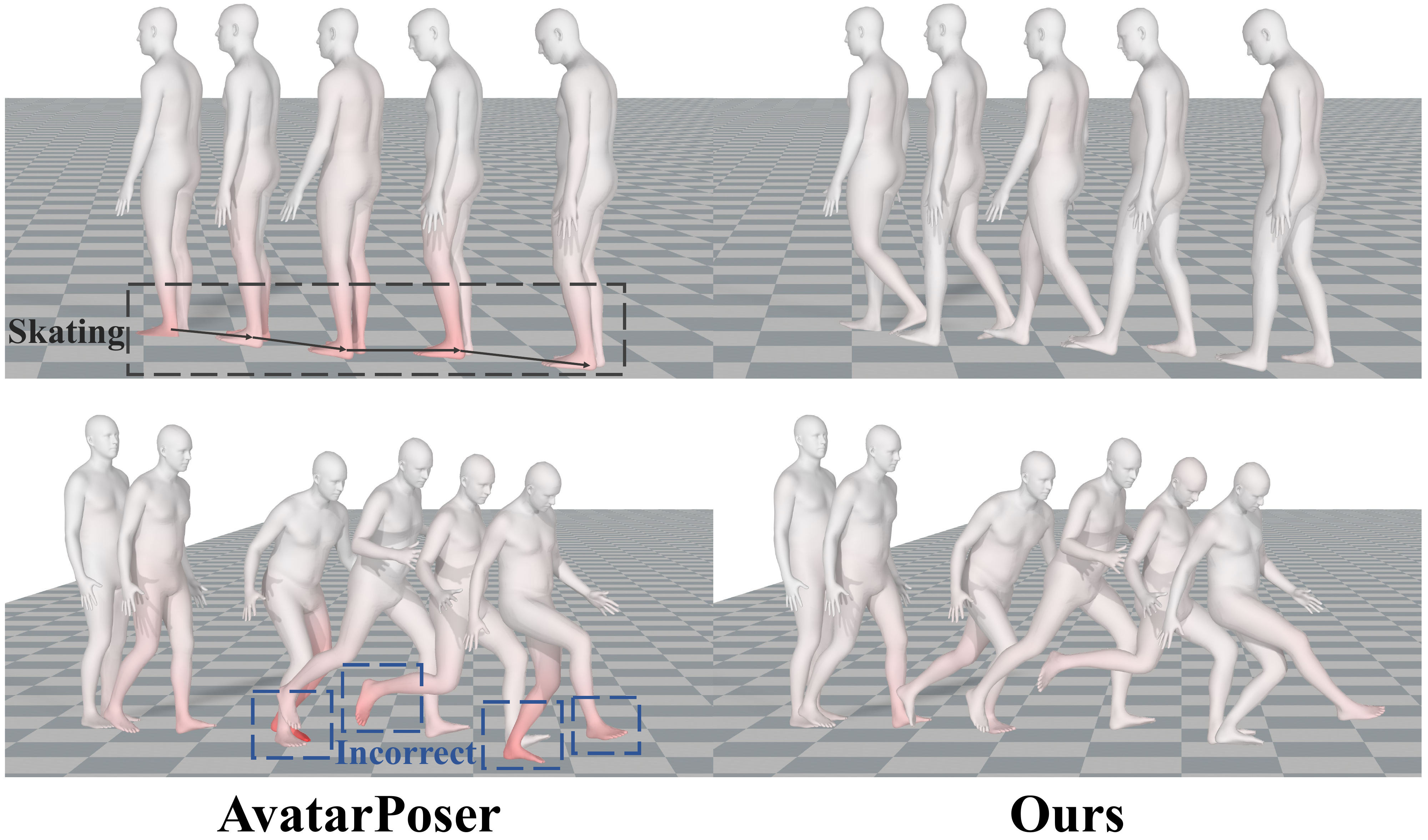}
    \vspace{-7pt} 
    \caption{Qualitative comparisons between AvatarPoser and ours. The results are color-coded to show errors in red.  
    }
    \label{fig:visual comparisons with ap} 
    \vspace{-19pt} 
\end{figure}

\section{Method}

\subsection{Problem Formulation}
In this task, we aim at predicting the full-body motion $\bm{\Theta} = \{\theta_{i}\}_{i=1}^{t} \in \mathbb{R}^{t \times s}$, given a sequence of sparse tracking signals $\bm{X} = \{x_{i}\}_{i=1}^{t} \in \mathbb{R}^{t \times c}$ of $t$ frames from headsets and hand controllers, where $c$ and $s$ denote the dimension of the input and output. 
Following the previous work \cite{jiang2022avatarposer}, we take the position, rotation, positional velocity, and angular velocity of the head and hands as the input signals; adopt pose parameters of the first 22 joints of the SMPL \cite{loper2015smpl} model to represent the output; and use the 6D representation of rotations for the input and SMPL model for its effectiveness \cite{zhou2019continuity}. 
Therefore, the input and output dimensions are $c=3 \times (6 + 6 + 3 + 3)=54$  and $s=22 \times 6=132$.

\begin{figure*}
    \centering
    \includegraphics[width=0.95\textwidth]{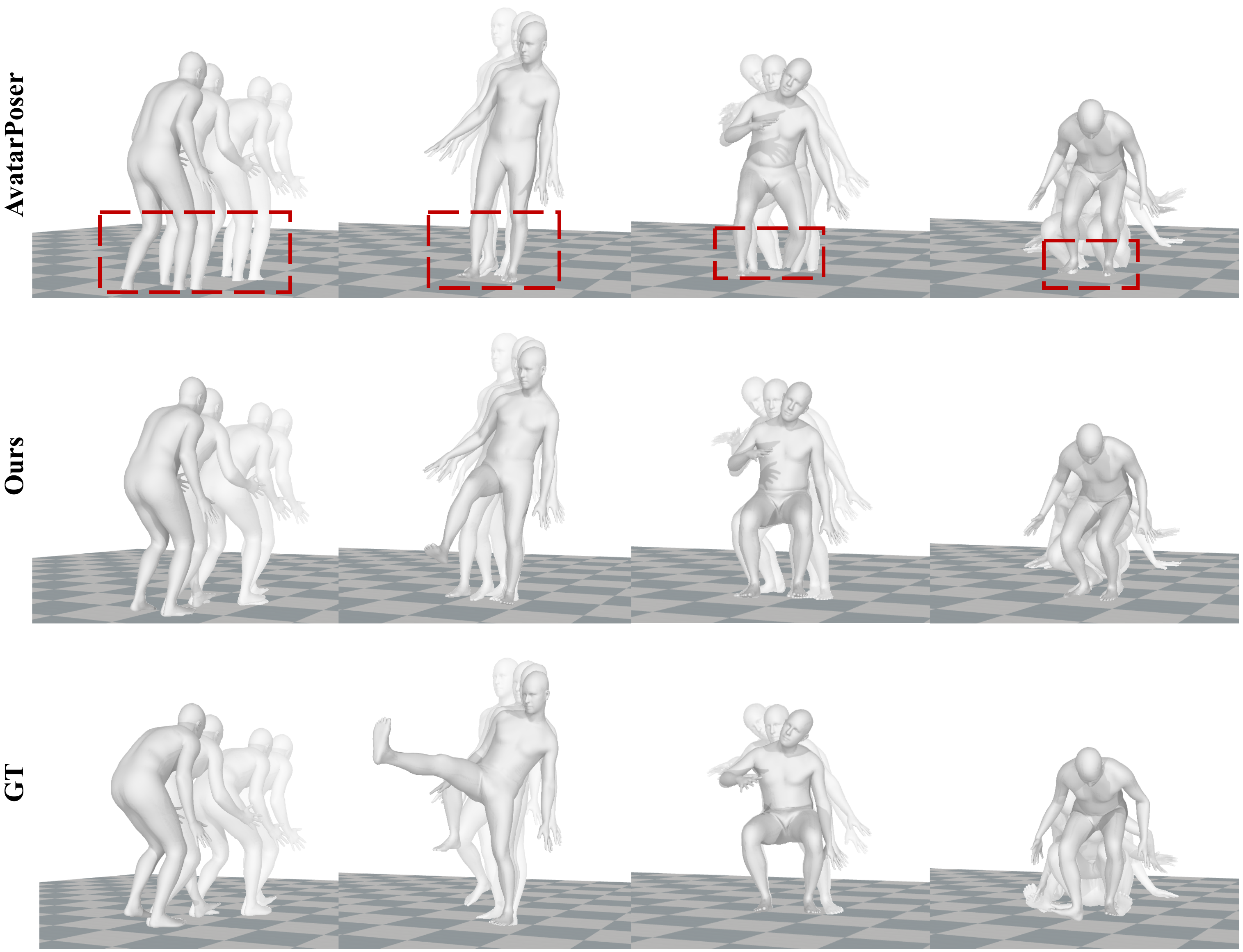}
    \vspace{-5pt} 
    \caption{Qualitative comparisons between AvataPoser, our method, and ground-truth on real-captured data.  }
    \label{fig:real-captured} 
    \vspace{-12pt} 
\end{figure*}

\vspace{-3pt} 
\subsection{Two-Stage Joint-Level Modeling Framework}
 As illustrated in \cref{fig:pipeline}, our joint-level modeling consists of two stages: 1) the initial joint-level features generation stage and 2) the joint-level spatiotemporal correlation modeling stage.
The first stage uses a simple network upon the input signals to generate the initial joint-level features serving as a basis for further exploiting the joint correlation. 
To achieve spatially accurate and temporally consistent full-body human motion, the second stage captures the joint dependencies in both spatial and temporal dimensions based on a spatiotemporal transformer-based network.

\subsubsection{Initial Joint-Level Features Generation}
To enable the explicit modeling of joint-level correlations, it is essential to generate initial joint-level features.
We accomplish this by 1) generating initial full-body poses and then 2) transforming them into joint-level positional and rotational features.
In this way, these two sorts of representations can explore different feature spaces to benefit following spatiotemporal modeling.

Specifically, the initial full-body pose for each frame is generated using two MLPs as follows:
\begin{equation}\label{eq:stage1}
\begin{split}
\bm{H}_{embed} = F_{embed}(\bm{X}) \in \mathbb{R}^{t \times d_1},\\
\bm{\Theta}_{init} = F_{reg}(\bm{H}_{embed})  \in \mathbb{R}^{t \times 22 \times 6},
\end{split}
\end{equation}
in which $F_{embed}$ embeds the input signals to high-dimensional input features $\bm{H}_{embed}$, $F_{reg}$ further regress the initial full-body pose $\bm{\Theta}_{init}$, and $d_{1}$ represents the dimension of the high-dimensional input features.

Next, we perform a forward-kinematics on the initial poses $\bm{\Theta}_{init}$ to obtain 3D positions of the joints $\bm{P}_{init} \in \mathbb{R}^{t \times 22 \times 3}$.
We subsequently convert these positions from the SMPL coordinate system to the head-relative coordinate system and convert the relative joint rotations to global joint rotations. 
Since we have specific observations of the head and hands, we substitute the positions and rotations of these joints with tracking signals, which provide guidance for the subsequent capture of joint correlations.
Then, we utilize two linear layers to embed the joint rotations and positions into high-dimensional joint-level features denoted as
$\bm{H}_{\Theta} \in \mathbb{R}^{t \times 22 \times d_2}$ and $\bm{H}_{P} \in \mathbb{R}^{t \times 22 \times d_2}$, respectively.
Here, $d_{2}$ is the dimension of the high-dimensional joint-level features.
Finally, we concatenate $\bm{H}_{\Theta}$ and $\bm{H}_{P}$ to obtain the initial joint-level features $\bm{H}_{init} = \left [\bm{H}_{\Theta} \otimes \bm{H}_{P}   \right ] \in \mathbb{R}^{t \times 44 \times d_2}$.
\subsubsection{Spatiotemporal Correlations Modeling}
To achieve spatially accurate and temporally consistent full-body human motion, we fully exploit the joint-level features in both spatial and temporal dimensions to model the spatiotemporal correlations.
Observed that our targeting full-body motion is dependent on the input signals from the head and hands, the most potential joint dependencies are related to the correlations between different joints and the head and hand joints.
To fully capture these correlations, we adopt a transformer-based network allowing easy long-range dependencies modeling between other joints and the head and hands joints. 
Therefore, the initial joint-level features are then utilized as carefully designed tokens for the spatiotemporal transformer-based network.

To fully excavate implicit information of designed tokens, we introduce a Spatial Transformer Block (STB) $F_{stb}$ and a Temporal Transformer Block (TTB) $F_{ttb}$ to capture the dependencies in the spatial and temporal dimensions, respectively. 
STB aims at capturing the spatial joint dependencies within a frame, especially the dependencies between other joints and observed head and hands joints, and achieving a reasonable single-frame pose estimation.
The input joint-level tokens 
$h_{i} \in \mathbb{R}^{44 \times d_{2}}$ 
include the joint rotational/positional features within a frame, where $h_{i}$ denotes the initial joint-level features $\bm{H}_{init}$ in the $i^{th}$ frame.
To preserve certain input information during the feature learning process, we supplement additional \emph{embedded input features} token $f_{i}$ to every single STB, where $f_{i}$ is the $i^{th}$ frame of $\bm{H}_{f} \in \mathbb{R}^{t \times 1 \times d_2}$ linear-transformed from $\bm{H}_{embed}$.
Therefore, the input tokens for STB are $s_{i}=\left [h_{i} \otimes f_{i}\right ] \in \mathbb{R}^{45 \times d_{2}}$
Note that before feeding $s_{i}$ into STB, we add a learnable spatial positional encoding $e_{s} \in \mathbb{R}^{45 \times d_{2}}$ to $s_{i}$ for indicating the relative location of the joint rotations/positions. 
Then, STB utilizes the self-attention mechanism \cite{vaswani2017attention} to model the dependencies of all the tokens for each frame and outputs the spatially enhanced joint-level features 
$\bm{H}_{s} = \{F_{stb}(s_{i})^{1:44}\}_{i=1}^{t}  \in \mathbb{R}^{t \times 44 \times d_{2}}$.

TTB focuses on learning the temporal correlations of each joint for maintaining temporal consistency and motion accuracy.
TTB treats each kind of feature across the sequence as tokens, resulting in the features $h^{i} \in \mathbb{R}^{t \times d_{2}}$ with $t$ tokens in total, where $h^{i}$ denotes the initial joint-level features $\bm{H}_{init}$ sliced in the second dimension. 
Besides, we add a learnable temporal positional encoding $e_{t} \in \mathbb{R}^{t \times d_{2}}$ to $h^{i} \in \mathbb{R}^{t \times d_{2}}$ for indicating the location of a specific joint feature in the sequence.
Then the temporally enhanced joint-level features $\bm{H_{t}} = \{F_{ttb}(h^{i})\}_{i=1}^{44}  \in \mathbb{R}^{t \times 44 \times d_{2}}$ are encoded by the TTB as outputs.

Spatial and temporal correlations modeling complement each other, in which STB tends to generate reasonable pose without temporal consistency and TTB tends to smooth the motion while introducing pose misalignment. 
Inspired by MixSTE~\cite{zhang2022mixste}, we alternatively use STB and TTB, which can decompose the feature learning into spatial and temporal dimensions.
Specifically, we stack STB and TTB for $n$ loops to obtain the final spatiotemporally modeled features $\bm{H_{st}} \in \mathbb{R}^{t \times 44 \times d_{2}}$.

Finally, we employ an MLP to regress the pose parameters $\bm{\Theta}$ of SMPL from $\bm{H_{st}}$. The MLP
consists of 2 linear layers, 1 group normalization layer, and 1 activation layer. The local full-body pose $\bm{P}$ is derived from $\bm{\Theta}$ using SMPL.

\begin{table}
  \centering
  \resizebox{0.99\columnwidth}{!}{
  \begin{tabular}{l|ccccccc}
    \toprule
    Input & Method & MPJRE & MPJPE & MPJVE  \\
    \midrule
    \multirow{6}*{Four}& Final IK      & 12.39 & 9.54 & 36.73      \\
    & CoolMoves     & 4.58  & 5.55 & 65.28      \\
    & LoBSTr        & 8.09  & 5.56 & 30.12      \\
    & VAE-HMD       & 3.12  & 3.51 & 28.23      \\
    & AvatarPoser   & 2.59  & 2.61 & 22.16  \\
    & Ours  & \textbf{2.40} & \textbf{2.09} & \textbf{17.82} \\
    \midrule
    \multirow{6}*{Three}
    & Final IK      
    & 16.77 & 18.09 & 59.24     \\
    & CoolMoves     
    &  5.20  & 7.83  & 100.54     \\
    & LoBSTr        
    &  10.69 & 9.02  & 44.97     \\
    & VAE-HMD       
    &  4.11  & 6.83  & 37.99     \\
    & AvatarPoser   
    & 3.21  & 4.18  & 29.40 \\
    & Ours          
    & \textbf{2.90}  & \textbf{3.35}  & \textbf{20.79}\\  
    \bottomrule
  \end{tabular}
  }
  \vspace{-9pt} 
  \caption{Evaluation results under Protocol 1. 
  }
  \label{tab:protocol 1.}
  \vspace{-5pt} 
\end{table}

\begin{table}
  \centering
  \resizebox{0.99\columnwidth}{!}{
  \begin{tabular}{l|lrrrr}
    \toprule
    Dataset & Method & MPJRE & MPJPE & MPJVE \\
    \midrule
    \multirow{6}*{CMU}
    & Final IK      
    & 17.80 & 18.82 & 56.83      \\
    & CoolMoves     
    & 9.20  & 18.77 & 139.17      \\
    & LoBSTr        
    & 12.51 & 12.96 & 49.94      \\
    & VAE-HMD       
    & 6.53  & 13.04 & 51.69      \\
    & AvatarPoser   
    & 5.93  & 8.37  & 35.76      \\
    & Ours  & \textbf{5.34}  & \textbf{7.75}  & \textbf{26.54} \\
    \midrule
    \multirow{6}*{BMLrub}
    & Final IK      & 15.93 & 17.58 & 60.64      \\
    & CoolMoves     & 7.93  & 13.30 & 134.77      \\
    & LoBSTr        & 10.79 & 11.00 & 60.74      \\
    & VAE-HMD       & 5.34  & 9.69  & 51.80      \\
    & AvatarPoser   
    & 4.92  & 7.04  & 43.70      \\
    & Ours   & \textbf{4.71}  & \textbf{6.49}  & \textbf{36.96} \\
    \midrule
    \multirow{6}*{HDM05}
    & Final IK      & 18.64 & 18.43 & 62.39      \\
    & CoolMoves     & 9.47  & 17.90 & 140.61      \\
    & LoBSTr        & 13.17 & 11.94 & 48.26      \\
    & VAE-HMD       & 6.45  & 10.21 & 40.07      \\
    & AvatarPoser   & 6.39  & 8.05  & 30.85      \\
    & Ours   & \textbf{5.86}  & \textbf{6.60}  & \textbf{23.57} \\

    \bottomrule
  \end{tabular}
  }
  \vspace{-9pt} 
  \caption{Evaluation results under Protocol 2. 
  }
  \label{tab:protocol 2}
  \vspace{-18pt} 
\end{table}

\begin{table}
  \centering
  \resizebox{0.99\columnwidth}{!}{
  \begin{tabular}{l|ccccccc}
    \toprule Method & MPJRE & MPJPE & MPJVE & Jitter  \\
    \midrule
     VPoser-HMD$^{\dagger}$     
     & /  & 6.74 & / & /      \\
     HuMoR-HMD $^{\dagger}$       
     & /  & 5.50 & / &  /    \\
     VAE-HMD$^{\dagger}$       
     & /  & 7.45 & / & /   
     \\
     ProHMR-HMD$^{\dagger}$       
     & /  & 5.22 & / &/   
     \\
     FLAG$^{\dagger}$      
     & /  & 4.96 & / &/   
     \\
     \midrule
     AvatarPoser$^{*}$   
     & 4.56 & 6.44 & 34.45 & 11.15  \\
     Ours   
     & \textbf{4.30} & \textbf{4.93} & \textbf{26.17} & \textbf{7.19} \\
    \bottomrule
  \end{tabular}
  }
  \vspace{-9pt} 
  \caption{Evaluation results under Protocol 3. $^{\dagger}$ denotes methods explicitly using additional pelvis information during inference. $^{*}$ denotes our retrained AvatarPoser using their public source code. 
  }
  \label{tab:protocol 3.}
  \vspace{-18pt} 
\end{table}

\begin{table*}
  \centering
  \resizebox{1.95\columnwidth}{!}{
  \begin{tabular}{c|ccccccccccc}
    \toprule
    Protocol & Method & MPJRE & MPJPE & MPJVE & Jitter & Ground & Skate & H-PE & U-PE & L-PE \\
    \midrule
    \multirow{2}*{1}
    & AvatarPoser$^{*}$       
    & 3.07 & 4.15 & 28.39 & 16.15 & 3.80 & 0.23 & 2.45 & 2.00 & 7.91 \\
    & Ours       
    & \textbf{2.90} & \textbf{3.35} & \textbf{20.79} & \textbf{8.39}  & \textbf{3.30} & \textbf{0.13} & \textbf{1.24} & \textbf{1.72} & \textbf{6.20} \\
    \midrule
    
    \multirow{2}*{3}
    & AvatarPoser$^{*}$     
    & 4.56 & 6.44 & 34.45 & 11.15& 2.95 & 0.32 & 3.70 & 2.93 & 12.59 \\
    & Ours
    & \textbf{4.30} & \textbf{4.93} & \textbf{26.17} & \textbf{7.19} & \textbf{2.17} & \textbf{0.21} & \textbf{1.45} & \textbf{2.27} & \textbf{9.59} \\
    \bottomrule
  \end{tabular}
  }
  \vspace{-9pt} 
  \caption{More metrics comparisons with AvatarPoser \cite{jiang2022avatarposer} under Protocol 1 and Protocol 3. $^{*}$ denotes our retrained AvatarPoser using their public source code. 
  }
  \label{tab:more metrics in protocol 1 and protocol 3.}
  \vspace{-5pt} 
\end{table*}

\begin{table*}
  \centering
  \begin{tabular}
  {l|ccccccccccc}
    \toprule
    Method & MPJRE & MPJPE & MPJVE & Jitter & Ground & Skate & H-PE & U-PE & L-PE  \\
    \midrule
    AvatarPoser$^{*}$  
    & 7.28 & 11.22 & 31.67 & 12.87 & 2.20 & 0.30 & 6.60 & 5.79 & 20.73  \\
    Ours
    & \textbf{6.98}
    & \textbf{9.52}
    & \textbf{25.78}
    & \textbf{10.04}
    & \textbf{0.20}
    & \textbf{0.21}
    & \textbf{5.31}
    & \textbf{5.16}
    & \textbf{17.15}
    \\    
    \bottomrule
  \end{tabular}
  \vspace{-9pt} 
  \caption{Evaluation results on the real-captured data. $^{*}$ denotes our retrained AvatarPoser using their public source code. }
  \label{tab: performance with real-captured data.}
  \vspace{-15pt} 
\end{table*}

\subsection{Loss Design and Training Process}
We make use of $L1$ body orientation loss, $L1$ body joint rotational loss, and $L1$ body joint positional loss, resembling AvatarPoser \cite{jiang2022avatarposer}.
Moreover, since this task is of a high degree of freedom, it is not easy to obtain accurate motion without additional constraints. Therefore, to better exploit the potential of our joint-level modeling, we introduce a set of losses that are tailored for this task to achieve better alignment with the input signals, temporal consistency, and physically plausible results.

\noindent
\textbf{Hand Alignment Loss.} As discussed in the previous study \cite{jiang2022avatarposer}, predicting absolute pelvis translations is worse than obtaining the translation from the known head position. Therefore, we also obtain the translation from 
\emph{head alignment} that aligns the head position of our local full-body pose to the global position of the head to obtain full-body positions 
in the global coordinate system $\bm{P}^{g}$. However, this approach causes a misalignment between the predicted and input global hand positions. To solve this problem, AvatarPoser adopts an inverse kinematics (IK) module to align hands. Nevertheless, the IK module is very slow. To better address this issue, we add a hand alignment loss $L_{h} = \frac{1}{2} \sum_{i=1}^{t}\|p_{i}^{hand} - \hat{p}_{i}^{hand}\|_{1}$
to align the global hands after the head alignment, where $p_{i}^{hand}$ denotes global hand joint positions of $i^{th}$ frame from $\bm{P}^{g}$ and $\hat{}$ denotes ground-truth.
This way makes the whole framework end-to-end trainable, contributing to good hand alignment without using the IK module to slow down the system. Since the full-body pose is highly correlated to the hand joints, our performance in different metrics all obtain improvement when the accurate hand position is perceived by the network. 

\noindent
\textbf{Motion Loss.} 
Following previous studies \cite{yi2021transpose,tevet2022human}, we also utilize velocity loss 
$L_{v}(l) = \frac{1}{t-1}  \sum_{i=1}^{t-1}\|(p_{i+l} - p_{i}) - (\hat{p}_{i+l} - \hat{p}_{i})\|_{1}$ and foot contact loss 
$L_{fc} = \frac{1}{t-1}  \sum_{i=1}^{t-1} \|(p^{feet}_{i+1} - p^{feet}_{i}) \cdot m_{i}\|_{1}$ for achieving a smooth and accurate motion, where $p_{i}$ is the predicted global full-body pose in $i^{th}$ frame and $p_{i}^{feet}$ denotes the joints relevant to the feet. 
$L_{v}$ encourages the inter-frame velocity to be close to the corresponding velocity of the ground-truth;
$L_{fc}$ enforces zero feet velocity when the feet are on the ground. $m_{i} \in \{0,1 \}^{k}$ is the binary foot contact mask of $i^{th}$ frame, denoting whether the feet touch the ground and $k$ is the feet-relevant joint number. Instead of only using $L_{v}(1)$ to supervise the velocity between adjacent frames, we also utilize $L_{v}(3)$ and $L_{v}(5)$ to avoid the accumulated velocity error. Therefore, the proposed motion loss is defined as :
\begin{equation}
    L_{mot} = L_{v}(1) + L_{v}(3) + L_{v}(5) + L_{fc}.
\end{equation}

\begin{figure}
    \centering
    \includegraphics[width=0.47\textwidth]{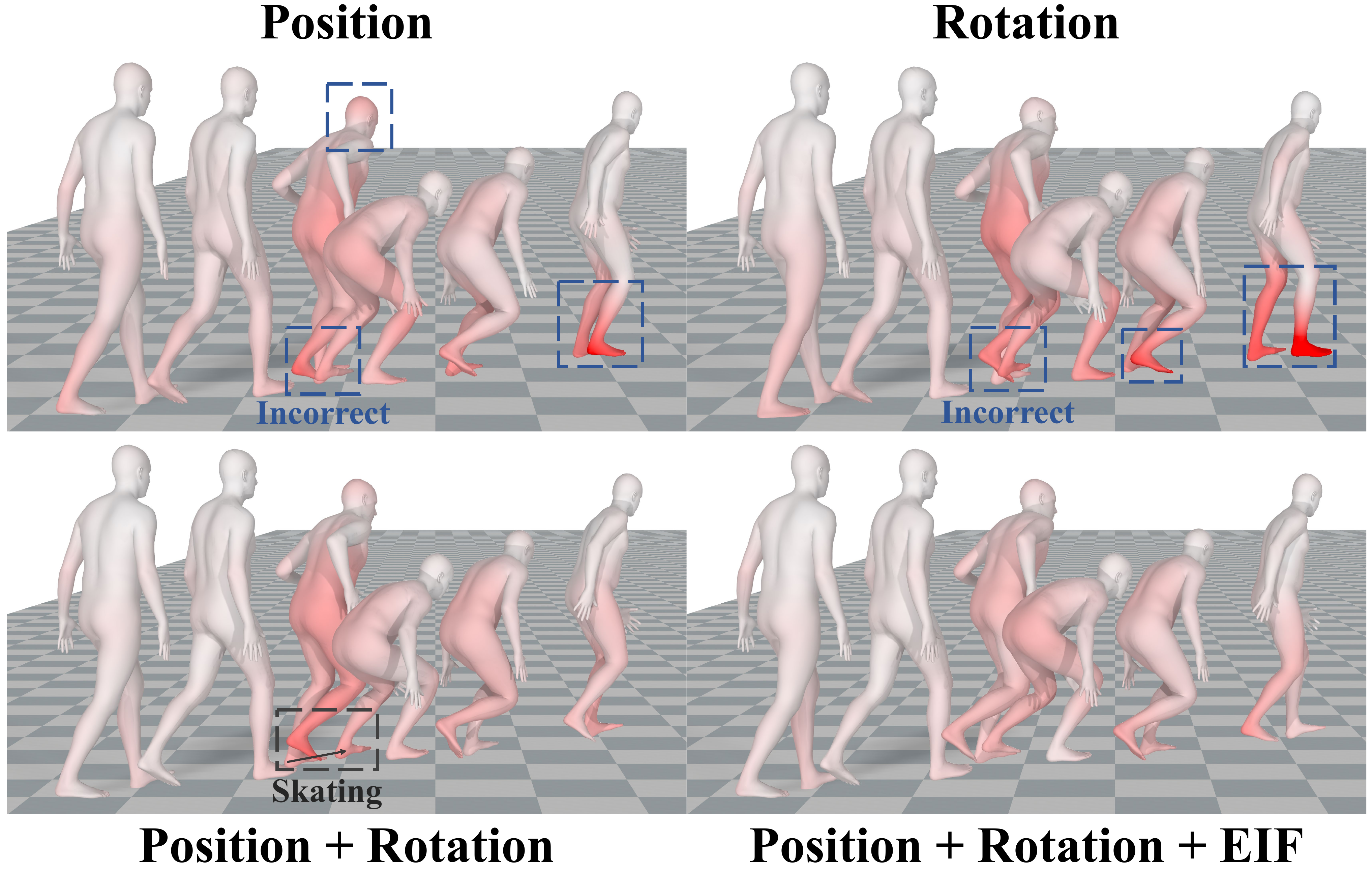}
    \vspace{-8pt}
    \caption{
    Ablation study for our method with four different generated features for the second stage, in which the errors are color-coded in red.
    }
    \label{fig:ablation initial feature.} 
  \vspace{-10pt}
\end{figure}

\noindent
\textbf{Physical Loss.}
Since this task lacks lower body information, the prediction tends to fail to match the upper body while maintaining the lower body physically plausible for challenging cases (e.g., jump, sit). To mitigate physical implausibility (especially ground penetration), we add ground penetration loss 
$L_{p} = \frac{1}{t}  \sum_{i=1}^{t} \|(z^{min}_{p_{i}} - z_{ground}) \cdot l_{i}  \|_{1}$ 
and ground-foot height loss
$L_{fh} = \frac{1}{t}  \sum_{i=1}^{t} \|(z^{feet}_{p_{i}} - z_{ground}) \cdot m_{i}  \|_{1}$,
where $z_{ground}$ is the ground height, $z^{min}_{p_{i}}$ is the predicted height of the lowest joint in $i_{th}$ frame, $z^{feet}_{p_{i}}$ is the predicted height of the feet-relevant joints in $i^{th}$ frame, and $l_{i}\in \{0,1 \}^{22}$ denotes whether the joint is lower than the ground. Thus, the proposed physical loss is defined as:
\begin{equation}
    L_{phy} = L_{p} + \alpha L_{fh}.
\end{equation}
These two physical losses are complementary to each other because penetration error tends to ``push'' the predictions away from the ground, while the foot height error tends to ``pull'' the predictions to the ground.

\noindent
\textbf{Overall Loss.}
Our complete loss function for training the model is defined as follows:

\begin{dmath}
    L = L_{first} + \beta L_{ori} + \gamma L_{rot} + \delta L_{pos} + \epsilon L_{h} + \zeta L_{mot} + L_{phy}, 
\end{dmath}
where $L_{first}$, $L_{ori}$, $L_{rot}$, and $L_{pos}$ are $L1$ loss for the initial head-relative full-body pose, the final SMPL root orientation $\bm{\Theta} ^{1:6}$, joint rotations $\bm{\Theta} ^{7:132}$, and full-body pose $\bm{P}$.
We set $\alpha$, $\beta$, $\gamma$, $\delta$, $\epsilon$, and $\zeta$ to 0.5, 0.02, 2, 5, 5, and 50 respectively to balance their loss scale.

\noindent
\textbf{Masked Training.}
During training, we randomly masked 2 of the tokens except for the head and hands tokens. In this way, the model is more robust for the quality of the generated initial joint-level features.

\noindent
\textbf{Implementation Detail.}
Following AvatarPoser \cite{jiang2022avatarposer}, we set the input sequence length $t$ to $41$ frames if not stated otherwise. 
During inference, we utilize sliding windows for prediction like AvatarPoser, which predicts the current frame using the current frame and previous $t-1$ frames.
We train the network with a batch size of 64 and use Adam solver \cite{kingma2013auto} for optimization. 
Our model is trained for $100,000$ iterations. 
The learning rate starts at $1e-4$ and drops to $1e-5$ after $60,000$ iterations. 
The feature dimension $d_{1}$ and $d_{2}$ for our model are set to $1024$ and $512$, and the stacked loops $n$ of our network are set to $6$.
Our model takes $24.7ms$ to infer $41$ frames on an NVIDIA A100 GPU without needing any further post-processing (e.g., IK).

\section{Experiment}
In this section, we first report the metrics adopted in previous tasks \cite{jiang2022avatarposer,aliakbarian2022flag,yuan2022physdiff,yi2021transpose}.
Then we compare with previous state-of-the-art methods qualitatively and quantitatively. 
We also evaluate each of our main contributions.
\cref{fig:real-captured} demonstrates the superiority of our methods using various motions. 
Please kindly refer to our supplementary video/materials for more details.

\noindent
\textbf{Accuracy-related metric.}
To evaluate the pose accuracy for each frame, we use \emph{MPJPE} (mean per joint position error) and \emph{MPJRE} (mean per joint rotation error), which measure the average position/relative rotation error of all body joints. Besides the full-body MPJPE, we also evaluate upper-body MPJPE (\emph{U-PE)}, lower-body MPJPE (\emph{L-PE}), and hand MPJPE (\emph{H-PE}), which can reflect different capabilities of the methods.

\begin{table*}
  \centering
  \resizebox{1.98\columnwidth}{!}{
  \begin{tabular}{c|lccccccccccc}
    \toprule
    Stage & Method & MPJRE & MPJPE & MPJVE & Jitter & Ground & Skate & H-PE & U-PE & L-PE \\
    \midrule
    \multirow{2} * {1} 
    & Constant Token   
    & 6.97 & 9.00  & 31.21 & 3.80 & 12.02 & 0.36 & 1.55 & 3.71 & 18.26 \\
    & Learnable Token 
    & 6.91 & 8.91  & 30.82 & 3.48 & 11.52 & 0.36 & 1.57 & 3.71 & 18.01 \\
    \midrule
    
    \multirow{4} * {2} 
    & Position            
    & 6.15 & 6.77  & 25.53 & 5.76 & 2.63 & 0.23 & \textbf{1.62} & \textbf{3.48} & 12.51 \\
    & Rotation            
    & 6.00 & 7.48  & 26.41 & \textbf{3.77} & \underline{2.48} & 0.26 & 1.92 & 3.62 & 14.21 \\
    & Rotation + Position         
    & \underline{5.90} & \underline{6.71}  & \underline{23.97} & 4.42 & 2.60 & \underline{0.22} & 1.71 & \underline{3.51} & \underline{12.30} \\
    & Rotation + Position + EIF      
    & \textbf{5.86} & \textbf{6.60}  & \textbf{23.57} & \underline{4.10} & \textbf{2.46} & \textbf{0.21} & \underline{1.69} & 3.52 & \textbf{12.12} \\
    \bottomrule
  \end{tabular}
  }
  \vspace{-9pt} 
  \caption{Performance comparisons between our proposed method with different initial joint-level features. The best results are in \textbf{bold}, and the second-best results are \underline{underlined}.}
  \label{tab:different initial joint-level features.}
  \vspace{-10pt}
\end{table*}

\begin{figure}
    \centering
    \includegraphics[width=0.47 \textwidth]{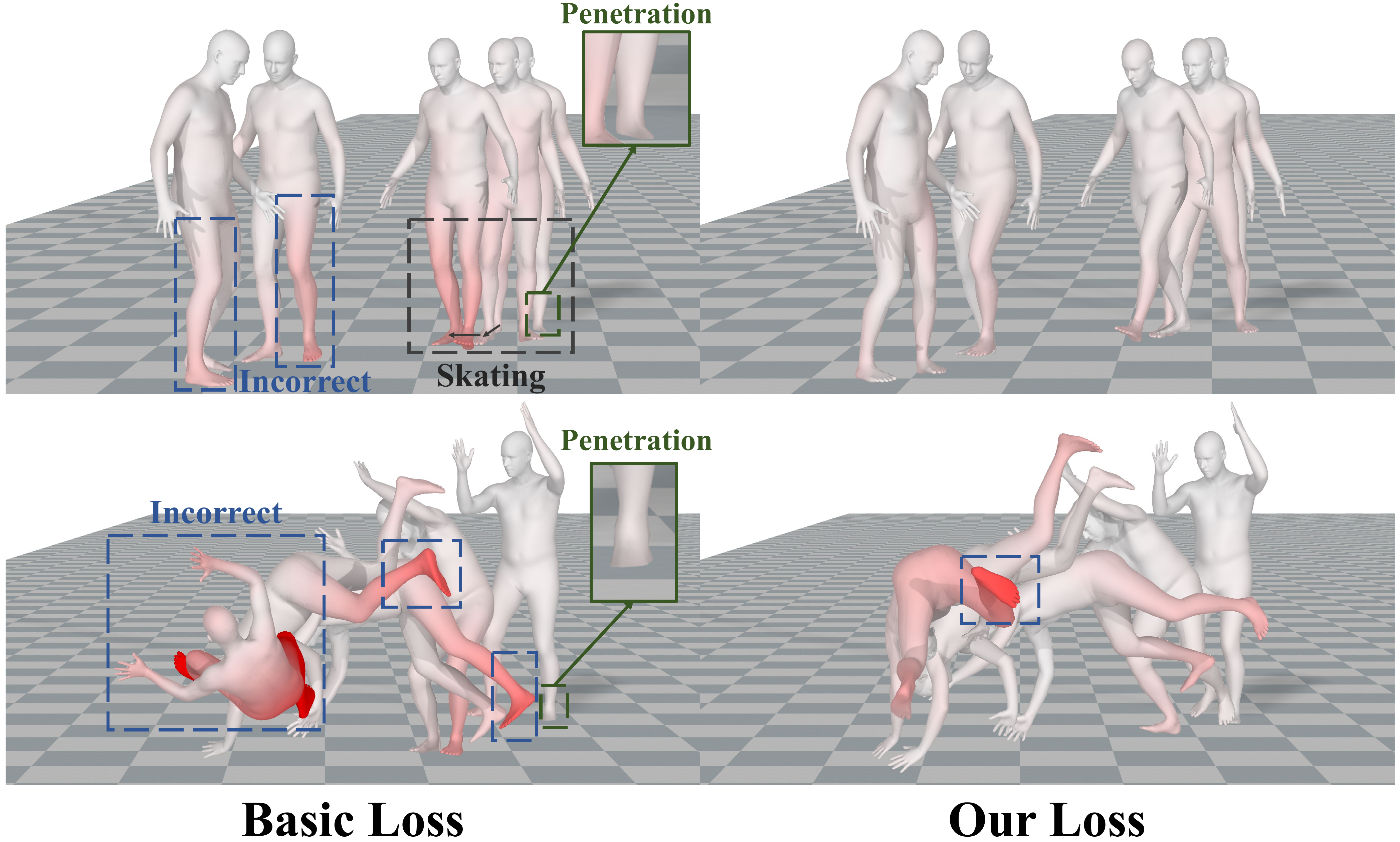}
    \vspace{-8pt}
    \caption{
    Ablation study for our method with different loss settings, in which the errors are color-coded in red.
    }
    \label{fig:ablation loss.} 
    \vspace{-12pt}
\end{figure}

\noindent
\textbf{Smoothness-related metric.}
To evaluate the motion smoothness, we use \emph{MPJVE} (mean per joint velocity error) and \emph{Jitter}. \emph{MPJVE} measures the average velocity error of all body joints. \emph{Jitter} measures the average jerk (time derivative of acceleration) of all body joints, which reflects the smoothness of the motion \cite{flash1985coordination}.

\noindent
\textbf{Physics-related metric.}
To evaluate the physical plausibility of motions, we use \emph{Ground} and \emph{Skate} metrics.
\emph{Ground} measures the distance between the lowest ground-truth body joint and the lowest predicted body joint. 
This metric reflects whether the generated motions have the correct contacting relationship with the ground (if the body penetrates the ground or floats above the ground, the error will be large).
\emph{Skate} measures the average horizontal displacement between the grounding feet in adjacent frames.

\begin{table*}
  \centering
  \resizebox{1.98\columnwidth}{!}{
  \begin{tabular}{l|ccccccccccc}
    \toprule
    Method & MPJRE & MPJPE & MPJVE & Jitter & Ground & Skate & H-PE & U-PE & L-PE \\
    \midrule
    Ours + Basic Loss
    & 6.09 & 7.50 & 32.53 & 8.98 & 3.66 & 0.35 & 3.11 & 3.93 & 13.76 \\
    + Hand 
    & 5.87 & 6.90 & 29.07 & 6.88 & 3.73 & 0.33 & \textbf{1.58} & \textbf{3.51} & 12.85 \\
    + Hand + Motion
    & \textbf{5.81} & \underline{6.74} & \underline{24.25} & \underline{4.22} & \underline{3.22} & \underline{0.22} & \underline{1.60} & 3.56 & \underline{12.32}  \\
    + Hand + Motion + Physical
    & \underline{5.86} & \textbf{6.60} & \textbf{23.57} & \textbf{4.10} & \textbf{2.46} & \textbf{0.21} & 1.69 & \underline{3.52} & \textbf{12.12}  \\
    \bottomrule
  \end{tabular}
  }
  \vspace{-9pt} 
  \caption{Performance comparisons between our proposed method with different loss functions. $^{*}$ denotes our retrained AvatarPoser using their public source code. The best results are in \textbf{bold}, and the second-best results are \underline{underlined}.
  }
  \label{tab:different losses.}
  \vspace{-12pt}
\end{table*}

\begin{table*}
  \centering
  \resizebox{1.98\columnwidth}{!}{
  \begin{tabular}{l|ccccccccccc}
    \toprule
    Method & MPJRE & MPJPE & MPJVE & Jitter & Ground & Skate & H-PE & U-PE & L-PE  \\
    \midrule
    AvatarPoser  
    & 6.39 & 8.05 & 30.85 & - & - & - & - & - & -  \\
    \midrule
    AvatarPoser-L$^{*}$  
    & 5.95 & 7.80 & 30.82 & 6.89 & 3.83 & 0.32 & 4.29 & 4.19 & 14.12  \\
    AvatarPoser-L$^{*}$ + Our Loss    
    & 6.02 & 7.14 & 23.92 & \textbf{3.41} & 2.51 & \textbf{0.21} & 1.82 & 3.54 & 13.43  \\    
    Ours + Our Loss    
    & \textbf{5.86} & \textbf{6.60} & \textbf{23.57} & 4.10 & \textbf{2.46} & \textbf{0.21} & \textbf{1.69} & \textbf{3.52} & \textbf{12.12}  \\
    \bottomrule
  \end{tabular}
  }
  \vspace{-9pt} 
  \caption{Architecture comparisons with AvatarPoser \cite{jiang2022avatarposer}. $^{*}$ denotes our retrained AvatarPoser using their public source code. AvatarPoser-L denotes a larger version of AvatarPoser. }
  \label{tab:model analysis.}
  \vspace{-12pt}
\end{table*}

\begin{table}
  \centering
  \resizebox{0.99\columnwidth}{!}{
  \begin{tabular}{l|cccccc}
    \toprule
    Method & MPJRE & MPJPE & MPJVE & Jitter \\
    \midrule
    Ours                   
    & \textbf{5.86} & \textbf{6.60} & \textbf{23.57} & \textbf{4.10} \\
    w/o Mask Training 
    & 5.91 & 6.65 & 23.87 & 4.19 \\
    \bottomrule
  \end{tabular}
  }
  \vspace{-9pt} 
  \caption{Performance comparisons between our proposed method with and without mask training.}
  \label{tab:mask training.}
  \vspace{-12pt}
\end{table}

\subsection{Comparison}
We make a thorough comparison with previous studies on the AMASS dataset\cite{mahmood2019amass}.
\cref{fig:visual comparisons with ap} presents the qualitative comparison against AvatarPoser \cite{jiang2022avatarposer} and demonstrates that our method can achieve more accurate and smooth results without physical implausibilities.
For quantitative comparison, we adopt three widely used settings as follows.

\noindent
\textbf{Protocol 1.} 
For the first setting, we follow \cite{jiang2022avatarposer} to split the subsets CMU \cite{cmu_amass}, BMLrub \cite{troje2002decomposing}, and HDM05 \cite{muller2007mocap} into 90\% training data and 10\% testing data. 
We compare our performance using both three (headset and controllers) and four inputs (add a pelvis tracker) with previous methods \cite{jiang2022avatarposer,dittadi2021full,ahuja2021coolmoves,yang2021lobstr,final_ik}. 
Note that unless otherwise stated, we all use three inputs for experiments.
As shown in \cref{tab:protocol 1.}, our performance with both three and four inputs outperforms existing approaches by a large margin in all previously used metrics. 
At the top of \cref{tab:more metrics in protocol 1 and protocol 3.}, we present comprehensive comparisons between our method and AvatarPoser using all the metrics. 
The results show that our method can accurately estimate lower body motions (\emph{L-PE}) while being much smoother and more physically plausible.

\noindent
\textbf{Protocol 2.}
For the second setting, we follow \cite{jiang2022avatarposer} to perform a 3-fold cross-dataset evaluation to compare with \cite{jiang2022avatarposer,dittadi2021full,ahuja2021coolmoves,yang2021lobstr,final_ik}. 
Using the subsets CMU \cite{cmu_amass}, BMLrub \cite{troje2002decomposing}, and HDM05 \cite{muller2007mocap}, we train on two subsets and test on the other subset in a round-robin fashion.
\cref{tab:protocol 2} shows the experimental results. 
Our method achieves the best performance over all the previously used metrics in all three datasets. 
Our performance is significantly better than the second-best method, demonstrating exploiting joint-level features contributes to generalization ability greatly.

\noindent
\textbf{Protocol 3.}
For the third setting, we follow \cite{aliakbarian2022flag} to use the subsets \cite{cmu_amass, akhter2015pose, trumble2017total, japanese_eye, mandery2015kit, troje2002decomposing, mandery2015kit, accad, mahmood2019amass, loper2014mosh, sfu, muller2007mocap} for training, and use the Transition \cite{mahmood2019amass} and HumanEva \cite{sigal2010humaneva} subsets for testing. 
The comparative results with \cite{pavlakos2019expressive,rempe2021humor,kolotouros2021probabilistic,dittadi2021full,jiang2022avatarposer} are shown in \cref{tab:protocol 3.}. 
Since some existing approaches \cite{pavlakos2019expressive,rempe2021humor,kolotouros2021probabilistic,dittadi2021full} implicitly assume the knowledge of the pelvis position but we do not, the comparisons are unfair. 
However, our \emph{MPJPE} is still superior to FLAG \cite{aliakbarian2022flag}, showing the effectiveness of our method. 
When fairly compared with AvatarPoser, our method is significantly better. 
The performance gap between AvatarPoser and ours is larger than that in other protocols, which indicates our method can benefit from more training data. 
At the bottom of \cref{tab:more metrics in protocol 1 and protocol 3.}, we present the comprehensive comparisons between our method and AvatarPoser using all the metrics. Similar to Protocol 1, our proposed method outperforms AvatarPoser in all metrics by a large margin.

Moreover, to further evaluate our performance on the real headset-and-controllers data for VR/AR applications.
We also capture a set of real evaluation samples from HMD and controller devices with the corresponding ground truth using a synchronized MoCap system; for details see supplementary materials.
To quantitatively compare our method with AvatarPoser and show the sim-to-real performance gap on real-captured data, we use the models trained with Protocol 3.
As shown in \cref{tab: performance with real-captured data.},  even though there are some acceptable drops in certain metrics from synthetic data, our model still outperforms AvatarPoser by a large margin as well, indicating our performance improvement does not come from simply overfitting the synthetic data but from well-learned motion knowledge from the large-scale motion data.
\cref{fig:real-captured} qualitatively demonstrates that our model can reconstruct more realistic and physically plausible results for those challenging cases (e.g., walking backward, kicking, sitting, and standing up) better than AvatarPoser.

Beyond the above comparisons, we also conducted a user study to compare the subjective quality of our method with AvatarPoser.  Our method achieved 3.69 scores while AvatarPoser gained 1.98 scores only~(5-level Likert scale). More details are in supplementary materials.

\subsection{Ablation Study}
We perform ablation studies using CMU \cite{cmu_amass} and BMLrub \cite{troje2002decomposing} for training and HDM05 \cite{muller2007mocap} for testing.

\noindent
\textbf{Initial joint-level features and tokens.}
To validate the effectiveness of our two-stage framework using a coarse full-body for feature initialization apart from the head and hand joints, we compare our method with two alternatives: 1) using the constant token as the initialized features; 2) using the learnable token as the initialized features, similar to \cite{carion2020end,ma2022ppt}. 
As shown in \cref{tab:different initial joint-level features.}, these two alternatives are much worse than ours.
Next, we demonstrate the effectiveness of our token design with different
initial joint-level features. 
The results in \cref{tab:different initial joint-level features.} proves that using either \emph{joint-rotation features} or \emph{joint-position features} is worse than combining them together to exploit different useful features. After adding \emph{embedded input features} token to provide input information for the spatial transformer blocks, the performance is better, indicating that repeatedly introducing the input information is useful for this task. 
\cref{fig:ablation initial feature.} shows visual results with different initial joint-level features.

\noindent
\textbf{Loss.}
\cref{tab:different losses.} shows the contribution of our designed loss functions. 
Adding hand alignment loss upon the basic loss significantly reduces hand error (\emph{H-PE}) and makes the network end-to-end trainable, which also improves other metrics.
Additional motion loss contributes a lot to the motion-related metrics (MPJVE, Jitter, and Sliding), and also benefit other metrics. 
Using physical loss, physics-related errors decrease, especially the \emph{Ground} error. 
For more experiments on every single combination of all our loss functions, please refer to our supplementary materials.
\cref{fig:ablation loss.} provides our visual results with and without our designed loss.
To better justify the effectiveness of our loss design, we also apply our complete loss function to the AvatarPoser. 
As shown in \cref{tab:model analysis.}, our loss can also dramatically improve AvatarPoser, indicating that our loss design is not only useful for our network design but also suitable for this task.

\noindent
\textbf{Architecture.}
To fully compare our joint-level architecture with AvatarPoser, we enlarge AvatarPoser by using 12 transformer layers and setting the feature dimension to 512 for it (if even larger, AvatarPoser cannot converge).
As shown in \cref{tab:model analysis.}, our method still outperforms AvatarPoser to a great extent.
Even after being significantly improved by our loss function, AvatarPoser still achieves an inferior performance to our method in almost all the metrics.
Even though the Jitter metric for AvatarPoser is better, analyzing this metric alone makes no sense since Jitter error can be quite low when predicting the same wrong full-body poses.
This phenomenon further justifies our joint-level modeling.

\noindent
\textbf{Mask training.}
\cref{tab:mask training.} shows that our method with and without mask training.
The results demonstrate that our method benefits from mask training. Meanwhile, this approach does not introduce any inference burden.

\begin{figure}
    \centering
    \includegraphics[width=0.46\textwidth]{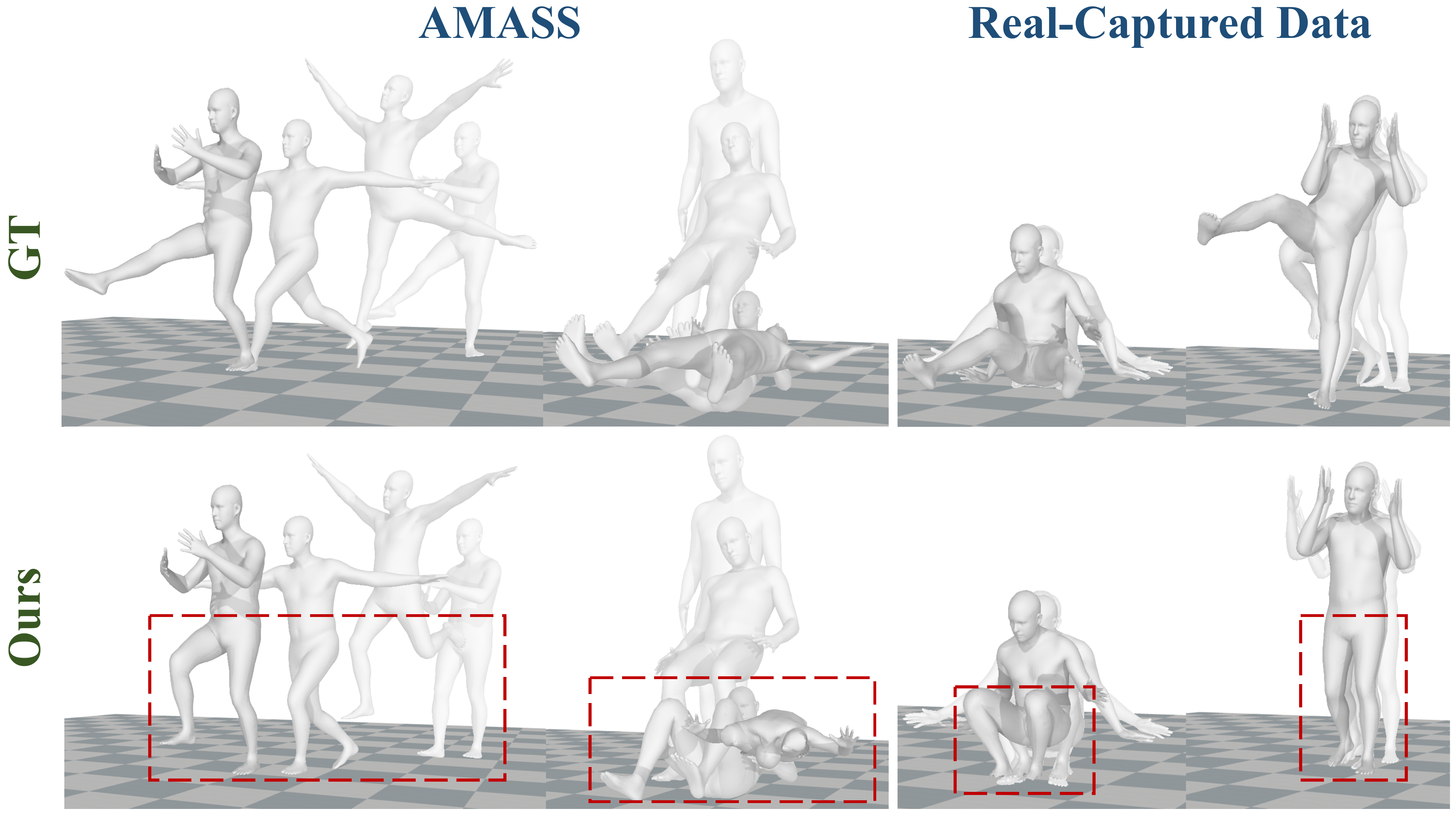}
    \vspace{-7pt} 
    \caption{Failure cases on AMASS and real-captured data.  }
    \label{fig:failure cases} 
    \vspace{-10pt} 
\end{figure}

\subsection{Limitation and Discussion}
We expect to generate realistic motions, in which ``realistic" emphasizes smoothness, physical plausibility, and alignment with the inputs. 
Despite our ability to achieve this objective in the majority of cases, there remain some issues that are beyond our capacity to solve.
\cref{fig:failure cases} presents our failure cases.
On AMASS, our model can not predict complex lower body motions (e.g., ballet) and may collapse with rare motions (e.g., falling down).
On real-captured data, our model has difficulty reconstructing motions with small variances of tracking signals but large lower-body movements.
Inferring full-body poses from three sparse signals is under-constrained and highly dependent on large-scale motion data.
Incorporating additional legs' signals (e.g., IMUs and images) may help resolve this problem. 
Furthermore, we believe more real-captured training data can also contribute greatly to the task since we observe some challenging cases can be solved with enough trained data on the synthetic data.

\section{Conclusion}
In this paper, we propose a two-stage learning-based framework for accurately estimating full-body motion from sparse head and hand tracking signals. 
We explicitly model the joint-level features in the first stage and then utilize them as spatiotemporal tokens for alternating spatial and temporal transformer blocks to estimate the full-body motion in the second stage.
With a set of carefully designed losses, we fully exploit the potential of our joint-level modeling to obtain realistic full-body motion.
Extensive experiments demonstrate both significant quantitative and qualitative improvement of our method over the previous state-of-the-art approaches without any post-processing during inference. 
We believe that our approach is a critical step in bridging the physical and virtual worlds for VR/AR applications.

\clearpage

\renewcommand{\thesection}{\Alph{section}}  
\renewcommand{\thetable}{\Alph{table}}  
\renewcommand{\thefigure}{\Alph{figure}}

\setcounter{section}{0}
\setcounter{figure}{0}
\setcounter{table}{0}

\twocolumn[{%
\begin{center}
    \textbf{\Large Supplementary Materials}
    \vspace{3em}
    \centering
    \captionsetup{type=figure}
    \includegraphics[width=0.95\textwidth ]{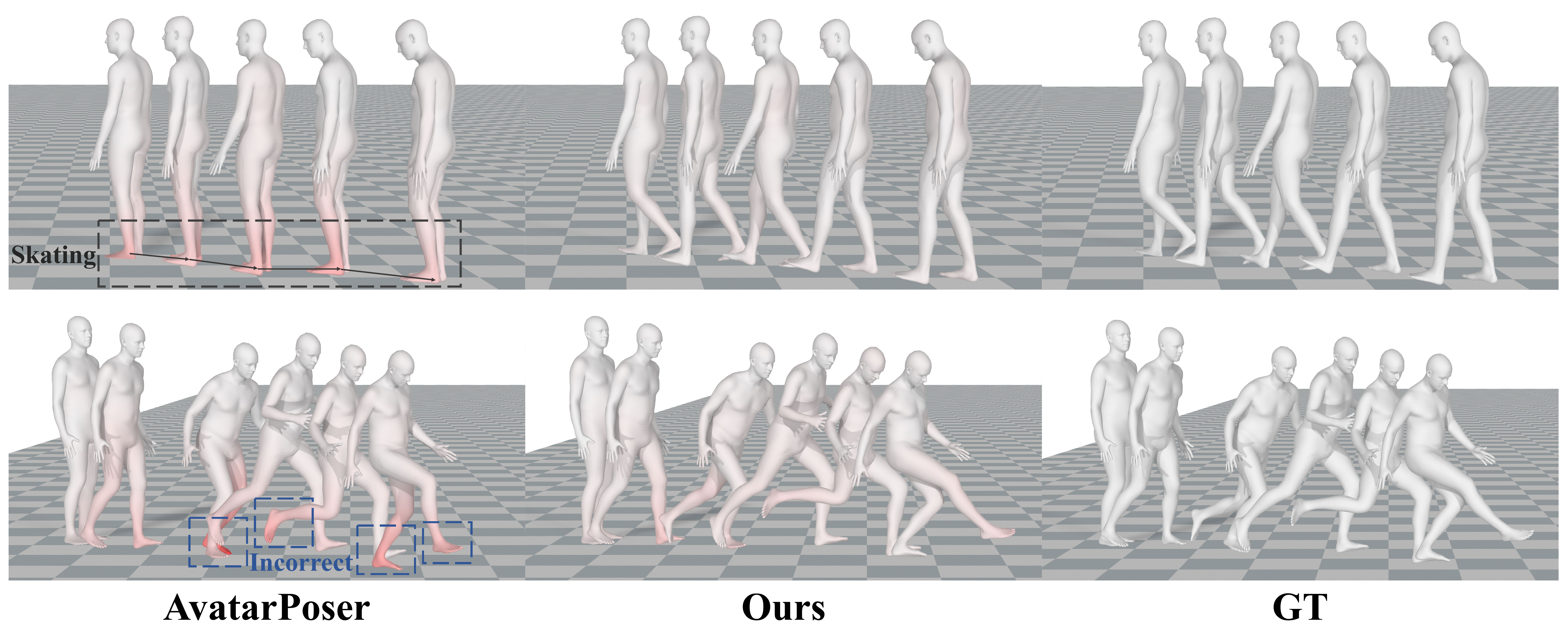}
        \captionof{figure}{Visual comparisons between various methods and ground truth.}
        \label{fig:visual comparisons with ap and GT}
\end{center}%
}]

\maketitle
\appendix

\section{Qualitative Comparisons}
\subsection{Error Color Coding}
For visualization, we use error color coding to show the difference between our reconstructed motion and GT.
The coded RGB value of a vertex is calculated as $[r,g,b]=(1-e)\times[204,204,204]+e\times[255,0,0]$, where $e$ is a vertex's error ($m$) clipped to 0$\sim$1.

\subsection{Comparison with GT}
To go beyond error color-coding in showcasing distinctions among various methods and Ground Truth (GT), we have also integrated GT into \cref{fig:visual comparisons with ap and GT}. 
The results illustrate that our model can capture realistic motions that are close to real motions. 
For more results, please refer to the supplementary video.

\subsection{Video Results}
The most effective way to qualitatively showcase motion tracking is through video results. Therefore, in addition to the highlighted figure results in the main manuscript, we also provide sequential qualitative comparison results for our proposed approach with AvatarPoser~\cite{jiang2022avatarposer}, both in the AMASS test set and real application scenarios. 
Please kindly refer to the accompanying supplementary video.

\subsection{User study.}
Based on the qualitative results, we also evaluate the subjective quality by conducting a user study. 
We randomly selected 25 participants from various schools and grades who were unfamiliar with the system. 
Participants rated the naturalness and realism of given motion sequences on a 5-level Likert scale. 
We computed mean opinion scores for each method, randomly selecting 12 motion samples in the test set and shuffling the list. Our method achieved 3.69 scores while AvatarPoser gained 1.98 scores only.

\section{Real-Captured Data}
To evaluate the model performance in real scenarios, we capture several sequences of real data with the corresponding ground truth.
The head and hands tracking signals are captured from the PICO 4 VR device, including both HMD and two controllers.
Besides, we also use a synchronized marker-based motion capture system, OptiTrack~\cite{optitrack}. Ground-truth SMPL parameters were then obtained from the MoCap data using MoSh++ \cite{mahmood2019amass}. 
To alleviate jittering, we apply the temporal filter to the ground-truth sequences.
For effective testing with real-world data, using the 6DoF (six degrees of freedom) inputs directly from the HMD and controllers might not be suitable. This is because practical wearing configurations can lead to gaps between the devices and the body joints.
To fill the sim-to-real gap, we apply empirically derived rigid transformations to convert the devices’ 6-DoF data to joint-based representations for evaluation. 
We will release our code with the above evaluation samples to facilitate future research in this field~\cite{projectpage}.

\section{Ablation Study}
Regarding the ablation study,
we also conduct additional experiments on transformer design, more loss combinations, and shorter input sequence lengths.

\subsection{Transformer Design}

\begin{table}
  \centering
  \begin{tabular}{cccccccccccc}
    \toprule
    Method  & MPJRE & MPJPE & MPJVE & Jitter  \\
    \midrule
    Ours  & 5.86 & 6.60 & 23.57 & 4.10  \\
    w/o STB & 6.03 & 7.19 & 24.34 & 4.21   \\
    w/o TTB & 6.13 & 7.30 & 27.99 & 5.14   \\
    \bottomrule
  \end{tabular}
  \caption{
  Performance comparisons between our proposed method with different transformer designs.
  }
  \label{table: ST}
\end{table}

As shown in \cref{table: ST}, removing STB or TTB both leads to significantly worse performance, indicating the importance of modeling spatial and temporal correlation simultaneously.

\begin{table*}
  \centering
  \resizebox{1.98\columnwidth}{!}{
  \begin{tabular}{l|ccccccccccc}
    
    \toprule
    Method & MPJRE & MPJPE & MPJVE & Jitter & Ground & Skate & H-PE & U-PE & L-PE \\
    \midrule
    Ours - Basic Loss
    & 6.09 & 7.50 & 32.53 & 8.98 & 3.66 & 0.35 & 3.11 & 3.93 & 13.76 \\
    \rowcolor{honeydew!200} 
    + Hand Alignment   
    & 5.87 & 6.90 & 29.07 & 6.88 & 3.73 & 0.33 & 1.58 & 3.51 & 12.85 \\
    \rowcolor{ghostwhite!200} 
    \quad + Velocity-Short    
    & 5.86 & 6.83 & 25.80 & 4.38 & 3.55 & 0.27 & 1.68 & 3.57 & 12.53 \\
    \rowcolor{ghostwhite!200} 
    \quad \quad + Velocity-Long     
    & 5.81 & 6.67 & 24.48 & 4.35 & 3.39 & 0.23 & 1.61 & 3.47 & 12.27 \\
    \rowcolor{ghostwhite!200} 
    \quad \quad \quad + Foot Contact 
    & 5.81 & 6.74 & 24.25 & 4.22 & 3.22 & 0.22 & 1.60 & 3.56 & 12.32 \\
    \rowcolor{aliceblue!200}
    \quad \quad \quad \quad + Penetration  
    & 5.85 & 6.80 & 23.34 & 3.87 & 2.65 & 0.20 & 1.69 & 3.50 & 12.58 \\
    \rowcolor{aliceblue!200}
    \quad \quad \quad \quad \quad  + Foot Height  
    & 5.86 & 6.60 & 23.57 & 4.10 & 2.46 & 0.21 & 1.69 & 3.52 & 12.12 \\
    \midrule
    \midrule
    + Hand 
    & 5.87 & 6.90 & 29.07 & 6.88 & 3.73 & 0.33 & 1.58 & 3.51 & 12.85 \\
    + Motion
    & 5.73 & 6.91 & 22.42 & 2.97 & 4.35 & 0.17 & 2.36 & 3.55 & 12.80  \\
    + Physical  
    & 5.99 & 8.30 & 33.24 & 8.54 & 2.63 & 0.31 & 5.12 & 4.61 & 14.75  \\ 
    + (Hand + Motion)
    & 5.81 & 6.74 & 24.25 & 4.22 & 3.22 & 0.22 & 1.60 & 3.56 & 12.32  \\
    + (Hand + Physical) 
    & 6.06 & 7.55 & 27.29 & 5.20 & 2.49 & 0.28 & 1.22 & 3.55 & 14.54  \\
    + (Motion + Physical) 
    & 5.90 & 7.18 & 22.93 & 3.26 & 2.43 & 0.17 & 2.41 & 3.73 & 13.22    \\
    + (Hand + Motion + Physical)
    & 5.86 & 6.60 & 23.57 & 4.10 & 2.46 & 0.21 & 1.69 & 3.52 & 12.12  \\
    \bottomrule
  \end{tabular}
  }
  \caption{
  Performance comparisons between our proposed method with different loss functions. 
  The different background color is used for indicating the category of the loss.
  \textcolor{honeydew!1000}{Green} denotes hand alignment loss; 
  \textcolor{ghostwhite!1000}{purple} denotes motion loss; 
  and \textcolor{aliceblue!1000}{blue} denotes physical loss.
  }
  \label{tab:different losses more.}
\end{table*}

\subsection{More Loss Combinations}
Our loss design substantially contributes to achieving accurate motion with temporal consistency.
Furthermore, it confers benefits not only to our method but also to other related approaches, such as AvatarPoser \cite{jiang2022avatarposer}, as indicated in the main manuscript.  
To better substantiate the efficacy of each individual loss term and assess the performance of different combinations of loss types, we conduct a comprehensive set of experiments. 
As detailed in the main manuscript, our total loss function consists of \emph{hand alignment} loss, \emph{motion} loss (velocity-short loss, velocity-long loss, and foot contact loss), and \emph{physical} loss (penetration loss and foot height loss).

We first add each loss term to the basic loss one by one.
The results and contribution of each individual loss term are presented at the top of Table \ref{tab:different losses more.}. 
As discussed in the main manuscript, the addition of hand alignment loss is crucial to render the entire framework end-to-end trainable and significantly enhances performance.  
Moreover, it better aligns the predicted motion with observed signals, resulting in more precise motion outputs.
The widely adopted velocity-short loss ($L_{v}(1)$) remarkably improves motion-related metrics, such as \emph{MPJVE} and \emph{Jitter}, as well as the \emph{Skate} metric. However, relying solely on $L_{v}(1)$ may not entirely eliminate accumulated velocity errors, necessitating the inclusion of additional velocity-long losses ($L_{v}(3)$ and $L_{v}(5)$). Our results indicate that velocity-long losses lead to further decreases in all error metrics, attesting to their effectiveness. 
Incorporating foot contact loss to constrain foot movement slightly enhances motion-related metrics. While the use of penetration loss can significantly reduce \emph{Ground} errors, it can also result in performance degradation in other evaluation metrics. This is because relying solely on penetration loss induces the network to predict results above the ground to reduce the \emph{Ground} error. Complementary foot-height loss can mitigate this issue, considerably reducing penetration errors while improving other evaluation metrics.

Subsequently, we explore the performance of various combinations of loss types, including \emph{hand alignment}, \emph{motion}, and \emph{physical} losses. 
The bottom of \cref{tab:different losses more.} demonstrates the performance of all different combinations.
Based on our experimental results, several conclusions can be drawn.
Firstly, it is crucial to frame the task as a sequence-to-sequence problem and employ the motion loss function accordingly.
Secondly, hand alignment loss is a complementary component that enhances the alignment of hands while simultaneously improving overall accuracy.
Thirdly, the physical loss term is a potent constraint that must be applied judiciously, as it can enhance performance only when the system already attains a substantially high level of accuracy and smoothness in motion.

\begin{table*}
  \centering
  \resizebox{1.98\columnwidth}{!}{
  \begin{tabular}{cccccccccccc}
    
    \toprule
    Method & Length & MPJRE & MPJPE & MPJVE & Jitter & Ground & Skate & H-PE & U-PE & L-PE \\
    \midrule
    AvatarPoser \cite{jiang2022avatarposer} & 41 & 3.21 & 4.18 & 29.40 & - & - & - & - & - & - \\ 
    \midrule
    Ours & 11 
    & 3.19 & 3.76 & 24.67 & 11.39 & 3.37 & 0.20 & 1.31 & 1.84 & 7.13 \\
    Ours & 21
    & 3.05 & 3.52 & 21.69 & 9.17 & 3.31 & 0.15 & 1.25 & 1.73 & 6.65  \\
    Ours & 41  
    & 2.90 & 3.35 & 20.79 & 8.39 & 3.30 & 0.13 & 1.24 & 1.72 & 6.20 \\
    \bottomrule
  \end{tabular}
  }
  \caption{
  Performance comparisons between our proposed method with different input sequence lengths. 
  }
  \label{tab:different lengths.}
\end{table*}
\label{suppsec:details}

\subsection{Shorter Input Sequence Length}
Although we have shown the possibility of applying our method in real scenarios in the attached video, achieving real-time performance for the application on mobile head-mounted displays (HMDs) with limited computing power is still challenging and important. Therefore, migrating to the HMDs is one of our future directions.

For exploring the possibility of our method applying to mobile devices, we perform experiments following Protocol 1 to evaluate our performance with shorter sequence lengths. 
The model's ability to process shorter input sequences is crucial for two reasons. 
First, it enhances the model's efficiency, which is essential for practical applications. 
Second, the capacity to handle short sequences enables the model to leverage future information with an acceptable latency.

According to the findings presented in \cref{tab:different lengths.}, our model exhibits robustness across varying sequence lengths. Even when processing very short sequences (11), our model demonstrates superior performance compared to AvatarPoser \cite{jiang2022avatarposer}. 
These results suggest that our model effectively leverages temporal information.

{\small
\bibliographystyle{ieee}
\bibliography{egbib}
}

\end{document}